%% file: main.tex
\pdfoutput=1

\documentclass{article} %
\usepackage{iclr2027_conference_arxiv,times}

\input{math_commands.tex}

\usepackage{amsmath,amssymb,amsthm}

\renewcommand{\eqref}[1]{\textup{(\ref{#1})}}
\usepackage{hyperref}
\usepackage{url}
\usepackage{amsmath,amssymb}
\usepackage{mathtools}
\usepackage{booktabs}
\usepackage[table]{xcolor}
\usepackage{float}      %
\usepackage{placeins}   %
\usepackage{enumitem}
\usepackage{tikz}
\usepackage{adjustbox}
\usepackage{bbm}
\usepackage{dsfont}
\usepackage{wrapfig}
\usepackage{multirow}
\usepackage{placeins}
\usepackage{placeins}
\usepackage{caption}
\usepackage{tabularx}
\usepackage{cuted}
\usepackage{xcolor}
\usepackage{soul}

\usetikzlibrary{arrows.meta,positioning}
\definecolor{oldgray}{RGB}{140,140,140}
\definecolor{newblue}{RGB}{0,70,180}
\definecolor{commentorange}{RGB}{220,120,0}
\usepackage{titlesec}
\titlespacing*{\section}{0pt}{6pt}{1pt}       %
\titlespacing*{\subsection}{0pt}{6pt}{1pt}    %
\titlespacing*{\subsubsection}{0pt}{6pt}{1pt} %

\newif\ifrevision
\revisionfalse           %

\ifrevision
    \newcommand{\oldtext}[1]{{\color{oldgray}#1}}
    \newcommand{\newtext}[1]{{\color{newblue}#1}}
    \newcommand{\duycomment}[1]{{\color{commentorange}#1}}
\else
    \newcommand{\oldtext}[1]{}
    \newcommand{\newtext}[1]{#1}
    \newcommand{\duycomment}[1]{}
\fi

\newtext{\title{\vspace{-0.25in}Retrieve to Act: Motion Primitive Graph Retrieval for Robust VLA Execution}}

\author{\textbf{Pham Tri Quang$^{1*}$, Bao-Ngoc Dao$^{1*}$, Minh Duc Nguyen$^{1,2}$, Duc-Duy Nguyen$^{1}$, Hung T. Ho$^{2}$,}\\[0.05in]
\textbf{Linh Dang Le$^{2}$, Trong-Bao Ho$^{2}$, Thien-Loc Ha$^{2}$, Tan Q. Nguyen$^{2}$, Tuan Dam$^{3}$, Vu N. Duong$^{1}$, }\\[0.05in] 
\textbf{Daniel Sonntag$^{4,5}$, Ngan Le$^{6}$, Khoa D Doan$^{1}$, Jan Peters$^{4,7}$, An Thai Le$^{1,2,7}$,}\\[0.05in]
\textbf{Duy M. H. Nguyen$^{4,8,9\dagger}$, Vien Anh Ngo$^{1,2\dagger}$}
\\[4pt]
$^{1}$Center for AI Research, VinUniversity\quad
$^{2}$VinRobotics\quad
$^{3}$Hanoi University of Science and Technology \\
$^{4}$German Research Center for Artificial Intelligence (DFKI)\quad
$^{5}$University of Oldenburg\\
$^{6}$University of Arkansas\quad
$^{7}$Technische Universit\"at Darmstadt\quad
$^{8}$Universit\"at Stuttgart \\
$^{9}$Max Planck Research School for Intelligent Systems (IMPRS-IS)\\
$^{*}$Equal Contributors \quad $^{\dagger}$Project Leads
}

\iclrfinalcopy %

\begin{document}

\maketitle

\vspace{-0.1in}
\begin{abstract}

\vspace{-0.1in}
\newtext{Vision-Language-Action (VLA) policies demonstrate remarkable semantic generalization, yet remain highly vulnerable to out-of-distribution (OOD) visual shifts during execution. While existing test-time methods attempt to suppress spurious visual cues, they fundamentally overlook the rich, uncorrupted kinematic priors embedded in the policy's own training experience. We introduce \textsc{Retrieve to Act (R2A)}, a test-time scaling framework that formulates OOD robustness purely as a geometric motion primitive retrieval problem. R2A bypasses fragile visual features by structuring successful training demonstrations into an offline spatiotemporal motion primitive graph. During inference, a query-dependent Graph Neural Network dynamically localizes the robot's ongoing execution within this graph, retrieving structurally compatible future motion primitives. These retrieved geometric priors are then seamlessly harmonized with the frozen VLA's predicted actions via a learned, dimension-wise gating mechanism, adaptively rectifying erroneous spatial trajectories while preserving the policy's local reactivity. Extensive experiments across three state-of-the-art VLA architectures, GR00T N1.7, $\pi_{0.5}$, and StableVLA, demonstrate that R2A consistently and significantly enhances robustness against diverse contextual perturbations. R2A achieves absolute success rate improvements of up to +12\% in simulation benchmarks (LIBERO-Plus, RoboTwin2.0-Plus) and +40\% in real-world deployments, without requiring any finetuning of the base policies.}

\textbf{Project page}: \url{https://retrieve2act.github.io/r2a/}
\end{abstract}

\section{Introduction}
\label{sec:intro}

Vision-Language-Action (VLA) models pretrained on large-scale datasets demonstrate remarkable semantic generalization, seamlessly interpreting novel instructions and adapting to diverse tasks~\citep{black2024pi0, intelligence2025pi_, bjorck2025gr00t}. However, this semantic capability often masks a severe kinematic fragility: even minor out-of-distribution (OOD) visual shifts, such as an altered camera angle, a novel background texture, or changed lighting, can cause catastrophic execution failures~\citep{fei2025libero, zhou2510libero, wang2025vlatest}. Under these conditions, the underlying task semantics remain identical, yet the policy's action generation diverges wildly from safe trajectories. This reveals a critical bottleneck for real-world deployment: while pretrained VLAs know \textit{what} to achieve, their over-reliance on perturbation-sensitive visual features causes them to easily lose track of \textit{how} to physically execute the task.

A natural approach to improving robustness is to finetune pretrained policies with augmented datasets that explicitly expose the models to perturbed conditions~\citep{xu2026apt, guo2026robustness}. However, exhaustively curating data and finetuning across the infinite long-tail of real-world environmental variations is both costly and practically impossible. Meanwhile, recent test-time scaling methods attempt to improve robustness without modifying model parameters by emphasizing task-relevant visual cues~\citep{hancock2025run, zhang2026probeact}, adapting via in-context information~\citep{jang2026ravla, sridhar2025ricl}, or steering action generation away from undesirable behaviors~\citep{liu2026vls, nguyen2026sdn}. While these approaches help mitigate the influence of visual noise or steer policies away from failure states, they fundamentally ignore a powerful asset: the policy's own successful training experience. The precise motion patterns, spatial relationships, and temporal transitions required to solve the task are already embedded in the training demonstrations. The challenge is that contextual perturbations obscure the correspondence between the current OOD observation and these learned motion primitives, preventing the policy from reusing its own prior successes.

Retrieval offers a natural mechanism to recall this prior experience, but applying flat retrieval to robotic trajectories is notoriously difficult. Prior robot-learning methods often retrieve demonstrations based on affordance similarity~\citep{kuang2024ram}, motion correspondence~\citep{lin2024flowretrieval}, or trajectory conditioning~\citep{park2026retrieve}. However, demonstrations are not isolated visual snapshots; they are highly structured spatiotemporal executions~\citep{xu2026stellavla}. Standard retrieval methods that match visual embeddings fail to capture this sequential geometry, conflating two distinct problems: \textit{where does the current execution lie within the trajectory space, and what specific motion primitive should follow?} To recover from an OOD-induced error, a retrieval mechanism must dynamically align the robot's ongoing geometric context with the underlying structure of prior demonstrations, entirely independent of visual noise.

To bridge this gap, we introduce \textbf{R2A} (\underline{R}etrieve to \underline{A}ct), a test-time scaling framework that formulates OOD robustness purely as a geometric experience retrieval problem. R2A bypasses fragile visual features by organizing successful training demonstrations into an offline \textit{motion primitive graph}. Nodes encode the pure kinematic geometry of past executions (poses, states, and frame-wise changes), while edges capture the spatiotemporal transitions across and within tasks. At the onset of an episode, a diffusion-based rectification module first projects any out-of-distribution initial robot states back into the established training manifold. During continuous execution, R2A then utilizes a \textit{query-dependent Graph Neural Network (GNN)} to dynamically propagate the VLA's current kinematic trajectory through the graph. This explicitly localizes the robot's current progress and retrieves the most structurally compatible future motion primitives. A learned harmonization module then adaptively weight-fuses this retrieved geometric prior with the VLA's predicted actions. By doing so, R2A seamlessly rectifies erroneous trajectories caused by visual OOD shifts, restoring successful motion patterns without altering the frozen VLA parameters.

We evaluate R2A across three diverse VLA architectures, including GR00T N1.7~\citep{bjorck2025gr00t}, $\pi_{0.5}$~\citep{intelligence2025pi_}, and StableVLA~\citep{fu2026stablevla}, under seven types of systematic perturbations in \texttt{LIBERO-Plus}~\citep{fei2025libero}. R2A consistently improves the robustness of all three policies, achieving gains of up to $+12\%$ in success rate. We further evaluate $\pi_{0.5}$ on the challenging \texttt{RoboTwin2.0-Plus} benchmark~\citep{zhang2026world} where R2A improves the success rate from $56.8\%$ to $66.2\%$. On a \texttt{real-world task with Aloha embodiment}, R2A similarly achieves up to a $40\%$ improvement in success rate over the corresponding base policies, demonstrating substantial gains in physical deployment.

\begin{figure*}[t]
  \centering
  \vspace{-0.2in}
  \includegraphics[width=0.9\textwidth]{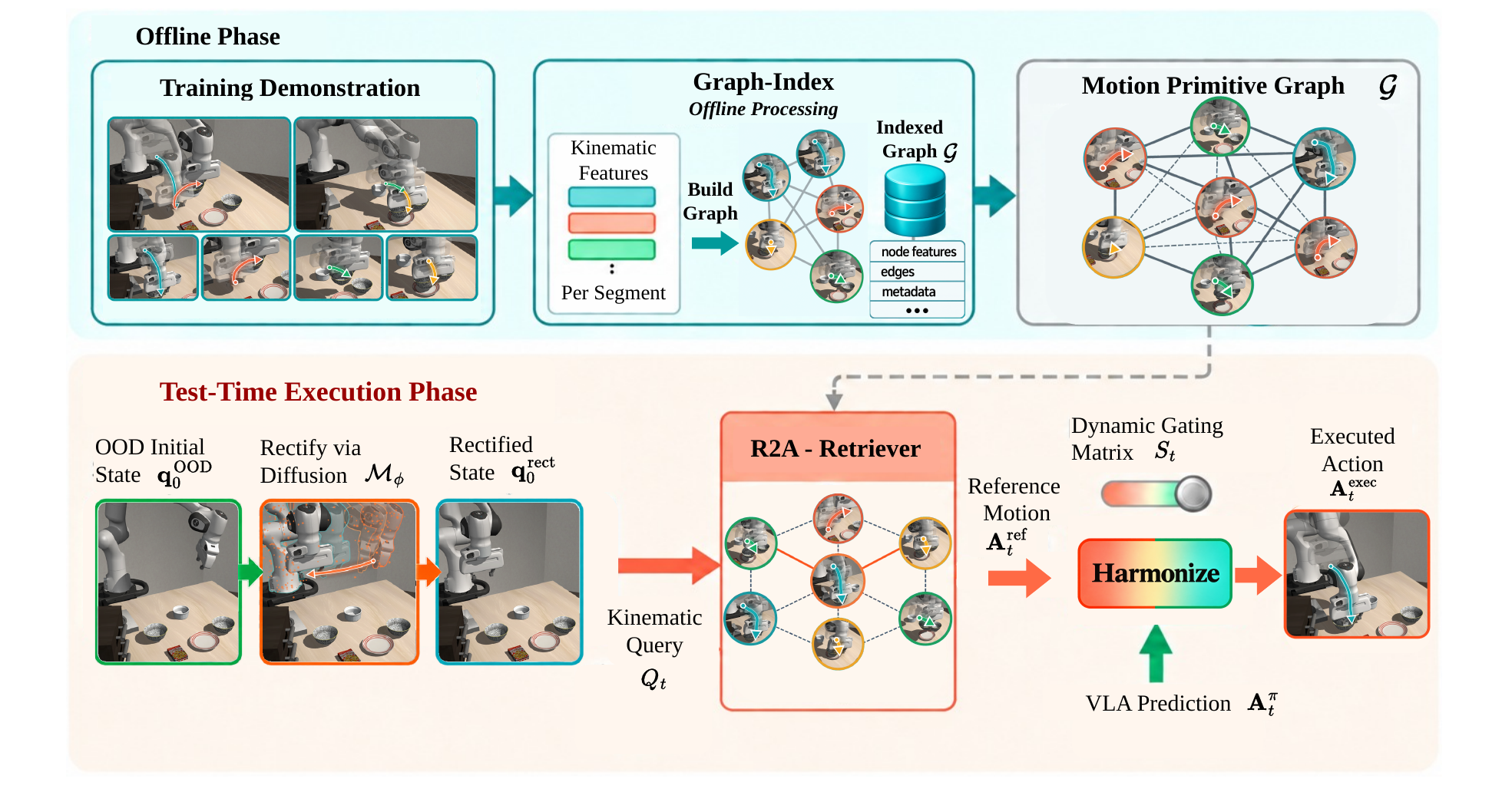}
  \vspace{-0.1in}
  \caption{\textbf{R2A} organizes offline training demonstrations into a pure kinematic motion primitive graph. At test time, it projects OOD initial states back to the training manifold via diffusion, continuously retrieves context-aligned geometric priors $\mathbf{A}_t^{\text{ref}}$ via a kinematic query $Q_t$, and harmonizes them with the frozen VLA’s predictions $\mathbf{A}_t^\pi$ via a dynamic gating matrix $\mathbf{S}_t$ to ensure robust execution under severe visual shifts. 
  }
  \label{fig:r2a_overview}
  \vspace{0.1in}
\end{figure*}
\section{Related Work} 
\paragraph{Test-Time Robustness of VLA Policies.}
Finetuning pretrained policies across the infinite long-tail of real-world environmental shifts is costly and practically impossible. Test-time scaling provides an efficient alternative, primarily by emphasizing task-relevant visual features or suppressing spurious cues at inference time. For instance, BYOVLA~\citep{hancock2025run} suppresses irrelevant regions via image inpainting, VLS~\citep{liu2026vls} relies on external vision models to generate steering rewards, and SDN~\citep{nguyen2026sdn} steers action generation away from OOD-induced negative behaviors. ProbeAct~\citep{zhang2026probeact} leverages 3D positional tracking to detect and recover from grasping failures.

Despite their effectiveness, these methods often require costly external visual modules~\citep{nguyen2026sdn, hancock2025run, liu2026vls} and over-rely on test-time visual cues, neglecting the rich behavioral priors already embedded in the training data. R2A overcomes this by structuring prior experience into a spatiotemporal motion primitive graph, bypassing fragile visual features entirely to dynamically retrieve and reinstate geometrically aligned behaviors.
\vspace{0.05in}
\vspace{-0.16in}
\paragraph{Retrieval-Augmented Robot Learning.}
Recent works demonstrate that useful experience need not be compressed entirely into policy parameters. Methods like RAM~\citep{kuang2024ram} construct reusable affordance memories, while Retrieve, Don't Retrain~\citep{park2026retrieve} conditions frozen policies on retrieved trajectories to avoid per-task finetuning. Other approaches emphasize low-level behavioral correspondence or action-sequence similarity to recover shared structures across visual variations~\citep{lin2024flowretrieval, lee2025class}.

While these works establish the value of external experience, they predominantly treat retrieval as a static matching problem between independent observations. Demonstrations, however, are highly structured temporal executions. R2A differentiates itself by treating the retrieval query not as an isolated snapshot, but as an \textit{ongoing geometric execution}. By mapping the robot's current trajectory into a structured graph of successful rollouts, R2A first dynamically localizes the execution state, and subsequently retrieves a structurally compatible future continuation.

\vspace{-0.15in}
\paragraph{Graph-Structured and Relational Retrieval.}

Graph-enhanced retrieval leverages structural dependencies to improve evidence selection, commonly using query-conditioned GNNs in semantic domains~\citep{luo2025gfmrag}. While recent robotics methods incorporate motion, they differ significantly in mechanism and scope. FlowRetrieval~\citep{lin2024flowretrieval} maps visual optical flow into a latent space for few-shot image retrieval, leaving it vulnerable to visual OOD shifts. CLASS~\citep{lee2025class} uses motion similarity strictly as a pretraining contrastive loss rather than for test-time retrieval. ReGIL~\citep{zhang2026regil} uses motion to generate auxiliary rewards for local trajectory alignment, but still fundamentally depends on standard, visually-driven retrieval pipelines.

R2A diverges by using motion to construct an explicit topological graph of \textit{pure kinematic states}, completely decoupled from fragile visual pixels. Rather than using motion for representation learning, visual flow matching, or auxiliary rewards, R2A deploys a query-conditioned GNN over this geometric graph to perform dynamic \textit{execution localization}. This explicitly spatial retrieval mechanism continuously infers the robot's ongoing position within prior structured successes, enabling robust retrieval of future geometric trajectories even under severe visual perturbations.

\vspace{-0.05in}
\section{Method} \label{sec:method}
\vspace{-0.05in}
\subsection{Preliminaries \& Problem Formulation}
\label{sec:preliminary}

At control step $t$, the robot's kinematic state comprises joint angles $\mathbf{q}_t \in \mathbb{R}^{7}$, the forward-kinematic end-effector pose $\mathbf{x}_t = \operatorname{FK}(\mathbf{q}_t) \in \mathbb{R}^{6}$, and a binary gripper state $\gamma_t \in \{0,1\}$. Conditioned on visual observations and a language instruction $\ell$, a frozen Vision-Language-Action (VLA) policy $\pi$ generates an action chunk $\mathbf{A}_t^\pi \in \mathbb{R}^{C\times7}$ every $C$ steps. We decouple this prediction into continuous spatial pose commands $\mathbf{A}_t^{\pi,\rm pose} \in \mathbb{R}^{C\times6}$ and discrete gripper commands $\mathbf{A}_t^{\pi,\rm grip} \in \mathbb{R}^{C\times1}$.

While $\pi$ generalizes well in-distribution, it remains highly sensitive to visual OOD shifts (e.g., altered viewpoints or lighting). These perturbations corrupt the visual observation space, frequently causing the spatial predictions $\mathbf{A}_t^{\pi,\rm pose}$ to diverge from safe, successful trajectories. To resolve this, we formulate OOD recovery as a visual-feature-free, geometry-grounded retrieval problem, dynamically correcting $\mathbf{A}_t^{\pi,\rm pose}$ using historical motion primitives from the training corpus.

\subsection{R2A Overview}
To explicitly leverage prior training experience, we introduce Retrieve to Act (R2A). Given a set of successful training demonstrations $\mathcal{D} = \{\tau_1, \tau_2, \dots, \tau_N\}$, we first construct an offline motion primitive graph $\mathcal{G} = (\mathcal{V}, \mathcal{E})$, where nodes $v \in \mathcal{V}$ encode the kinematics of demonstration segments (poses, states, and frame-wise changes) and edges $e \in \mathcal{E}$ capture spatiotemporal geometry.
During test-time execution, R2A (i) rectifies out-of-distribution initializations, (ii) queries the graph to retrieve geometrically relevant motion primitives, and (iii) harmonizes them with the policy's original prediction. The overall R2A pipeline is formulated as:
\begin{equation*}
\adjustbox{max width=\linewidth}{%
\begin{tikzpicture}[node distance=4mm,
  box/.style={draw, rounded corners=2pt, line width=0.5pt, inner xsep=3pt, inner ysep=3pt},
  flow/.style={-{Stealth[length=4pt]}, line width=0.6pt}]
  \node[box] (rect) {$\mathbf{q}_0^{\rm rect} = \text{Rectify}(\mathbf{q}_0^{\rm OOD}, \mathcal{M}_\phi)$};
  \node[box, right=of rect] (ret) {$\mathbf{A}_t^{\rm ref} = \text{R2A-Retriever}(Q_t, \mathcal{G})$};
  \node[box, right=of ret] (harm) {$\mathbf{A}_t^{\rm exec} = \text{Harmonize}(\mathbf{A}_t^{\rm ref}, \mathbf{A}_t^\pi)$};
  \node[box, above=4mm of ret] (graph) {$\mathcal{G} = \text{Graph-Index}(\mathcal{D})$};
  \draw[flow] (rect) -- (ret);
  \draw[flow] (ret) -- (harm);
  \draw[flow] (graph) -- (ret);
\end{tikzpicture}}
\vspace{-2mm}
\end{equation*}

In this framework (Fig.~\ref{fig:r2a_overview}), $\text{Graph-Index}(\cdot)$ processes the training demonstrations into $\mathcal{G}$. At the start of an episode, $\text{Rectify}(\cdot)$ projects any out-of-distribution initial robot states back into the training manifold using a diffusion model $\mathcal{M}_\phi$. During continuous execution, the $\text{R2A-Retriever}$ constructs a kinematic query $Q_t$ by combining the VLA's current prediction with the robot's recent state history. To localize the most relevant experiences, we formulate a query-dependent GNN~\citep{zhu2021neural,galkin2024towards} that dynamically conditions the message-passing process on the active query $Q_t$, outputting a robust reference motion $\mathbf{A}_t^{\rm ref}$. Finally, the $\text{Harmonize}(\cdot)$ module adaptively fuses $\mathbf{A}_t^{\rm ref}$ with the VLA's prediction to generate the final executed action $\mathbf{A}_t^{\rm exec}$, dynamically correcting trajectory deviations while preserving the policy's local reactivity.

\subsection{Motion Primitive Graph Construction} \label{sec:graph}
The first step in the R2A pipeline is processing the offline training demonstrations $\mathcal{D}$ into a structured motion primitive graph $\mathcal{G} = (\mathcal{V}, \mathcal{E})$, corresponding to the $\text{Graph-Index}(\mathcal{D})$ operation. Unlike prior methods that index visual embeddings, we construct $\mathcal{G}$ entirely based on the spatiotemporal geometry of the robot's execution trajectories.

\textbf{Node Representation}. We extract continuous segments from the $N$ demonstrations to form the node set $\mathcal{V}$. To bypass visual OOD shifts, we strictly discard visual features and initialize each node $v \in \mathcal{V}$ by encoding its kinematic geometry and temporal order $\displaystyle \mathbf{h}_v^{(0)} = \bigl[ \operatorname{norm}(\mathbf{q}_v, \delta\mathbf{q}_v, \gamma_v) \,\Vert{}\, \operatorname{PE}(\bar{s}_v) \bigr],$ where $[\cdot \,\Vert{}\, \cdot]$ denotes concatenation and $\operatorname{norm}(\cdot)$ normalizes the kinematic states. Here, $\mathbf{q}_v \in \mathbb{R}^7$ is the joint configuration, $\delta\mathbf{q}_v$ is the joint displacement across the segment, and $\gamma_v \in \{0,1\}$ is the gripper state. To retain sequential context, $\operatorname{PE}(\bar{s}_v)$ applies a positional encoding based on the node's relative temporal position $\bar{s}_v$ within its demonstration. Finally, each node is augmented with two metadata attributes for retrieval: a semantic task label $y_v$ and an estimated progress score $p_v \in [0,1]$ quantifying the temporal distance to task completion.

\textbf{Edge Construction}. We define the edge set $\mathcal{E} = \mathcal{E}_{\text{temp}} \cup \mathcal{E}_{\text{spatial}}$ to capture both sequential motion dynamics and cross-task geometric similarities:
\begin{itemize}[itemsep=0pt, topsep=2pt, leftmargin=15pt, parsep=3pt]
    \item \textbf{Temporal Edges} ($\mathcal{E}_{\text{temp}}$): We connect each node at time step $t$ to near-future nodes at $t+\Delta$ ($\Delta \in \{2, 4, 6, 8\}$) within the same demonstration, embedding a strong forward-looking trajectory prior.
    \item \textbf{Spatial Edges} ($\mathcal{E}_{\text{spatial}}$): To enable cross-context behavioral sharing, we connect any two nodes $v_i, v_j$ across the entire dataset if the geometric distance between their end-effector poses falls below a predefined threshold. This creates topological shortcuts between kinematically aligned states, regardless of their source task.
\end{itemize}

\subsection{Initial State Rectification via Diffusion}
A critical vulnerability of VLA policies is their sensitivity to out-of-distribution (OOD) initializations~\citep{fei2025libero}; when the robot's starting arm configuration drifts significantly from the training demonstrations, the policy's visual-proprioceptive grounding breaks down before execution begins. To counteract this, R2A employs a diffusion-based state rectification step at $t=0$ to project the OOD initial state back into the established training distribution. 

We model the distribution of valid starting configurations using a conditional diffusion model $\mathcal{M}_\phi$. During training, we sample valid initial joint configurations $\mathbf{q}_0$ from the demonstration dataset $\mathcal{D}$. We corrupt $\mathbf{q}_0$ by adding Gaussian noise $\epsilon \sim \mathcal{N}(\mathbf{0}, \mathbf{I})$ over $K$ forward steps to produce a noisy state $\mathbf{q}^k$. The denoising network $\epsilon_\phi$ is trained to predict the injected noise, conditioned on the current diffusion step $k$ and the semantic task instruction $\ell$. The training objective is the standard mean squared error noise-prediction loss:
\begin{equation}
    \mathcal{L}_{\text{Diff}}(\phi) = \mathbb{E}_{\mathbf{q}_0 \sim \mathcal{D}, \epsilon \sim \mathcal{N}(\mathbf{0}, \mathbf{I}), k} \left[ \Vert{} \epsilon - \epsilon_\phi(\mathbf{q}^k, k, \ell) \Vert{}_2^2 \right].
\end{equation}
During test-time inference, if the robot initializes in an OOD configuration $\mathbf{q}_0^{\rm OOD}$, we utilize the trained model to sample a valid in-distribution starting configuration. Starting from pure Gaussian noise $\mathbf{q}^K \sim \mathcal{N}(\mathbf{0}, \mathbf{I})$, the model iteratively denoises the sample conditioned on the task instruction $\ell$ as $\mathbf{q}^{k-1} = \mathbf{q}^k - \epsilon_\phi(\mathbf{q}^k, k, \ell), \quad \text{for } k = K, \dots, 1.$.

Once the reverse process yields a valid target state $\mathbf{q}_0^{\rm rect} = \mathbf{q}^0$, we compute a rigid corrective trajectory to physically transition the robot from $\mathbf{q}_0^{\rm OOD}$ to $\mathbf{q}_0^{\rm rect}$. By rectifying the initial state prior to engaging the VLA, we ensure that subsequent kinematic queries $Q_t$ fed into the GNN retriever remain geometrically grounded within the bounds of the motion primitive graph.

\subsection{Query-Dependent GNN Retrieval}
Once the motion primitive graph $\mathcal{G}$ is constructed, R2A formulates trajectory recovery as a dynamic retrieval process. The $\text{R2A-Retriever}$ localizes the most relevant historical experiences by propagating a kinematic query through a query-dependent GNN, followed by a marginalization and re-ranking step to isolate the top-$K$ nodes.

\textbf{Query Initialization}. At control step $t$, we first construct a kinematic query $Q_t$ that encapsulates both the VLA's intended future motion and the robot's recent execution context. Specifically, we concatenate the VLA's current pose prediction
$\mathbf{A}_t^{\pi,\rm pose}$ with the recent history of robot states $\mathbf{x}_{t-H:t}$ and executed actions $\mathbf{A}_{t-H:t}^{\rm exec}$, and process the entire sequence through an encoder network:$$Q_t = \text{Encoder}\Bigl(\bigl[ \mathbf{A}_t^{\pi,\rm pose} \,\Vert{}\, \mathbf{x}_{t-H:t} \,\Vert{}\, \mathbf{A}_{t-H:t}^{\rm exec} \bigr]\Bigr),$$
where $H$ denotes the history window size.

\vspace{0.05in}
\textbf{Query-Dependent Message Passing.}
Unlike conventional GNNs~\citep{gilmer2017neural} that perform static feature aggregation, our query-dependent GNN conditions the message-passing operations directly on $Q_t$. During each GNN layer $l$, the representation $\mathbf{h}_v^{(l)}$ for node $v$ is updated by aggregating messages from its spatial and temporal neighbors $\mathcal{N}(v)$, where both the message computation and the node update are dynamically modulated by the active query $Q_t$:
\begin{equation*} 
\mathbf{m}_{u \to v}^{(l)} = \text{Message}^{(l)}\bigl(\mathbf{h}_u^{(l-1)}, \mathbf{h}_v^{(l-1)}, Q_t\bigr); \ \mathbf{h}_v^{(l)} = \text{Update}^{(l)}\Bigl(\mathbf{h}_v^{(l-1)}, \text{Aggregate}_{u \in \mathcal{N}(v)} \bigl(\mathbf{m}_{u \to v}^{(l)}\bigr), Q_t\Bigr). 
\label{eq:gnn}
\end{equation*}

By explicitly conditioning the network on $Q_t$, the GNN actively propagates query relevance across the graph's geometric topology. After $L$ layers of message passing, the network outputs a relevance probability distribution over all nodes in the graph: $w_v = \text{Softmax}\bigl( \text{MLP}(\mathbf{h}_v^{(L)}, Q_t) \bigr) \quad \forall v \in \mathcal{V},$where $w_v \in [0,1]$ indicates the initial relevance of node $v$ to the ongoing rollout.

\vspace{0.1in}
\textbf{Marginalization and Context Estimation}. The relevance scores $\{w_v\}_{v \in \mathcal{V}}$ induce a discrete probability distribution over the graph conditioned on the query $Q_t$. To infer the macroscopic execution context, we marginalize over this distribution to compute the expected values of the node attributes. Let $\mathbf{y}_v$ denote the one-hot task label and $p_v \in [0,1]$ denote the temporal progress scalar for node $v$. The expected task distribution $\hat{\mathbf{y}}_t$ and expected trajectory progress $\hat{p}_t$ are formally evaluated as:
\begin{equation*} \hat{\mathbf{y}}_t = \mathbb{E}_{v \sim \mathbf{w}}[\mathbf{y}_v] = \sum_{v \in \mathcal{V}} w_v \mathbf{y}_v,;\  \hat{p}_t = \mathbb{E}_{v \sim \mathbf{w}}[p_v] = \sum_{v \in \mathcal{V}} w_v p_v. 
\end{equation*}
These expectations ground the active query, providing a statistical estimate of which task the robot is currently attempting and how close it is to completion.

\vspace{0.1in}
\textbf{Re-ranking and Top-K Selection.} 
We incorporate the marginalized context estimators to update the initial query representation, producing a context-aware query $\tilde{Q}_t = \text{MLP}\bigl([Q_t \,\Vert{}\, \hat{\mathbf{y}}_t \,\Vert{}\, \hat{p}_t]\bigr)$. The final retrieval score $s(v)$ for each node $v \in \mathcal{V}$ is computed via cosine similarity between the updated query and the terminal node representations, augmented by a bias term reflecting the alignment of the task and progress estimates:
   $\displaystyle s(v) = {\tilde{Q}_t \cdot \mathbf{h}_v^{(L)}/}{\Vert{}\tilde{Q}_t\Vert{} \Vert{}\mathbf{h}_v^{(L)}\Vert{}} + \text{Bias}(\mathbf{y}_v, p_v, \hat{\mathbf{y}}_t, \hat{p}_t).$
To isolate the most relevant geometric experiences, we extract the Top-$K$ node set $\mathcal{V}_K \subset \mathcal{V}$, strictly defined such that $\vert{}\mathcal{V}_K\vert{} = K$ and $\min_{v \in \mathcal{V}_K} s(v) \geq \max_{u \in \mathcal{V} \setminus \mathcal{V}_K} s(u)$. This operation is compactly denoted as: $\mathcal{V}_K = \operatorname*{TopK}_{v \in \mathcal{V}} \, s(v).$ Let $\mathbf{A}_v^{\text{demo}} \in \mathbb{R}^{C \times 6}$ denote the recorded future pose commands associated with the demonstration segment corresponding to node $v$. The reference motion chunk $\mathbf{A}_t^{\rm ref}$ is generated via a normalized score-weighted fusion over the retrieved set: $\displaystyle {\mathbf{A}_t^{\rm ref} = \sum_{v \in \mathcal{V}_K} \alpha_v \mathbf{A}_v^{\text{demo}}, \quad \text{where} \quad \alpha_v = {\exp(s(v)/\tau)}/{\sum_{u \in \mathcal{V}_K} \exp(s(u)/\tau)},}$  with $\tau$ acting as a temperature hyperparameter to control the sharpness of the fusion. This aggregated reference motion is subsequently passed to the $\text{Harmonize}(\cdot)$ module.

\vspace{0.1in}
\textbf{Training Objective.}
To train the query-dependent GNN to output accurate relevance scores $w_v$, we employ a ranking-based contrastive learning paradigm. During training, we sample an anchor node $u$ from a demonstration and formulate a corresponding training query $Q_u$. 

\textit{- Positive and Negative Mining:} We define a positive target set $\mathcal{P}_u$ for the anchor $u$ by filtering its local graph neighborhood. Specifically, $\mathcal{P}_u$ includes temporal neighbors that are close in task progress (e.g., immediate next steps) and spatial neighbors whose geometric feature distance from $u$ falls below a strict threshold. The negative set $\mathcal{N}_u$ comprises the remaining nodes connected to $u$ that fail to satisfy these temporal or spatial proximity conditions.

\textit{- Loss Formulation:} To maximize the likelihood of retrieving these geometrically relevant nodes, we first compute the binary cross-entropy (BCE) loss:
\begin{equation}
    \mathcal{L}_{\text{BCE}} = - \frac{1}{\vert{}\mathcal{P}_u\vert{}} \sum_{v \in \mathcal{P}_u} \log w_v - \frac{1}{\vert{}\mathcal{N}_u\vert{}} \sum_{v \in \mathcal{N}_u} \log(1 - w_v).
\end{equation}
However, due to the high sparsity of valid target nodes within the broader graph, relying solely on BCE can lead to vanishing gradients~\citep{lin2024understanding}. To mitigate this and explicitly maximize the margin between positive and negative nodes, we introduce an auxiliary ranking loss~\citep{bai2023regression}: 
\begin{equation}
    \mathcal{L}_{\text{RANK}} = - \frac{1}{\vert{}\mathcal{P}_u\vert{}} \sum_{v \in \mathcal{P}_u} \frac{w_v}{\sum_{v' \in \mathcal{N}_u} w_{v'}}.
\end{equation}
The final training objective for the GNN is optimized via a weighted combination of the BCE and ranking losses:
$\mathcal{L} = \alpha \mathcal{L}_{\text{BCE}} + (1 - \alpha)\mathcal{L}_{\text{RANK}},$ where $\alpha \in (0, 1)$ is a scaling hyperparameter. By minimizing $\mathcal{L}$, the network learns to dynamically assign high relevance scores $w_v$ to topologically and geometrically aligned experiences while suppressing distant, irrelevant nodes.

\section{Action Harmonization} 
To preserve local reactivity while rectifying out-of-distribution (OOD) errors, R2A harmonizes the geometrically robust reference motion $\mathbf{A}_t^{\rm ref}$ with the VLA's prediction via a learned, dimension-wise gating matrix $\mathbf{S}_t \in [0, 1]^{C \times 6}$: $\displaystyle \mathbf{S}_t = \sigma\Bigl( \text{MLP}\bigl( [ Q_t \,\Vert{}\, \mathbf{A}_t^{\pi,\rm pose} \,\Vert{}\, \mathbf{A}_t^{\rm ref} ] \bigr) \Bigr).$Conditioned on the discrepancy between the predicted and reference motions, $\mathbf{S}_t$ implicitly detects OOD perturbations. The final executed action applies this gate as a residual correction to spatial poses, leaving discrete gripper commands unaltered:$$\mathbf{A}_t^{\rm exec,pose} = \mathbf{A}_t^{\pi,\rm pose} + \mathbf{S}_t \odot \bigl(\mathbf{A}_t^{\rm ref}-\mathbf{A}_t^{\pi,\rm pose}\bigr); \quad \mathbf{A}_t^{\rm exec,grip} = \mathbf{A}_t^{\pi,\rm grip},$$where $\odot$ denotes the Hadamard product. This ensures R2A intervenes selectively, restoring successful motion primitives only when the VLA's spatial actions deviate from the geometric prior.\vspace{0.1in}

\textbf{Harmonization Training}. The R2A framework is trained sequentially: we first optimize the base VLA policy $\pi$, then the query-dependent GNN, and finally the harmonization MLP. The MLP is trained while keeping the pretrained VLA and GNN frozen. To ensure robust OOD recovery rather than mere in-distribution memorization, we simulate perceptual perturbations by injecting noise $\epsilon$ into the kinematic query, yielding $\tilde{Q}_t = Q_t + \epsilon$. With MLP parameters $\theta$, the network minimizes the mean squared error against ground-truth demonstration actions $\mathbf{A}_t^{\text{gt,pose}}$:
\begin{equation}
    \mathcal{L}_{\text{Harmonize}}(\theta) = \mathbb{E}_{t \sim \mathcal{D}, \epsilon} \left[ \left\Vert{} \mathbf{A}_t^{\pi,\rm pose} + \mathbf{S}_t(\tilde{Q}_t; \theta) \odot \bigl(\mathbf{A}_t^{\rm ref} - \mathbf{A}_t^{\pi,\rm pose}\bigr) - \mathbf{A}_t^{\text{gt,pose}} \right\Vert{}_2^2 \right].
\end{equation}
This objective explicitly forces the model to increase the gating weights in $\mathbf{S}_t$ and rely on the retrieved geometric prior $\mathbf{A}_t^{\rm ref}$ whenever simulated noise compromises the VLA prediction.

\input{section/experiment}

\input{section/conclusion}
\section*{Reproducibility Statement}
We provide detailed experimental settings, training procedures, models, datasets, evaluation protocols, and parameters in Appendix~\ref{app:implement} to facilitate reproducibility. Our code is available at an anonymous repository: \url{https://anonymous.4open.science/r/r2a}.
\section*{AI Use Statement}
We used large language models (LLMs) solely to improve the grammar, clarity, and readability of the manuscript. All research ideas, methods, experiments, analyses, and conclusions are the authors' own, and the authors reviewed and take full responsibility for all content of this paper.

\input{main.bbl}
\clearpage
\appendix
\input{appendix/appendix}
\input{appendix/implementation_detail}

\end{document}

%% file: math_commands.tex
\usepackage{amsmath,amsfonts,bm}

\def\eqref#1{equation~\ref{#1}}

\def\1{\bm{1}}

\DeclareMathAlphabet{\mathsfit}{\encodingdefault}{\sfdefault}{m}{sl}
\SetMathAlphabet{\mathsfit}{bold}{\encodingdefault}{\sfdefault}{bx}{n}

%% file: section/experiment.tex
\section{Experiments}
\label{sec:experiments}
\vspace{-0.05in}
\paragraph{Main Results.}
We conduct experiments in both simulation and real-world environments to evaluate whether R2A improves frozen VLAs under deployment-time perturbations. Implementation details are provided in Appendix~\ref{app:implement}. To systematically evaluate R2A, we design our experiments around the following research questions:

\input{table/main_table}

\textbf{Q1. Does R2A improve the robustness of existing VLA policies?}
We evaluate whether R2A improves the robustness of pretrained VLA policies under execution-context perturbations. We augment three representative VLAs, GR00T N1.7~\citep{bjorck2025gr00t, nvidia2026gr00t17}, $\pi_\text{0.5}$~\citep{intelligence2025pi_}, and StableVLA~\citep{fu2026stablevla}, with R2A and evaluate them across seven perturbation categories: Camera Viewpoints, Robot Initial States, Language Instructions, Light Conditions, Background Textures, Sensor Noise, and Object Layouts. Our baselines include state-of-the-art VLAs: OpenVLA~\citep{kim2024openvla}, OpenVLA-OFT~\cite{kim2025fine}, NORA~\citep{hung2025nora}, WorldVLA~\citep{cen2506worldvla}, UniVLA~\citep{bu2025univla}, $\pi_{0}$-FAST~\citep{pertsch2025fast}, RIPT-VLA~\citep{tan2505interactive} and VLA-JEPA~\citep{sun2026vla}.

As shown in Table~\ref{tab:main}, R2A consistently improves all three base policies, increasing the average success rate from 77.1\% to 89.0\% for GR00T N1.7, 78.1\% to 89.5\% for StableVLA, and 85.9\% to 90.4\% for $\pi_\text{0.5}$. The R2A-enhanced policies also outperform recent strong VLA policies such as VLA-JEPA~\citep{sun2026vla}, StableVLA~\citep{fu2026stablevla}, and $\pi_\text{0.5}$~\citep{intelligence2025pi_}, demonstrating the effectiveness of motion-primitive retrieval for improving robustness without modifying the underlying policies.
\input{table/ood_and_real}

Across perturbation categories, R2A improves $\pi_\text{0.5}$ in six of seven categories, with only a marginal decrease under Background Textures, while maintaining comparable performance under Light Conditions. StableVLA improves consistently across all seven categories. The largest gains occur under Robot Initial States, where R2A improves GR00T N1.7 by 49.1\%, StableVLA by 48.0\%, and $\pi_\text{0.5}$ by 21.9\%. This highlights the benefit of retrieving motion primitives with relevant spatial and temporal context, particularly when the robot starts from states substantially different from those encountered during pretraining.

\input{table/r2a-demo}

\vspace{0.05in}
We further evaluate R2A on bimanual manipulation in RoboTwin2.0-Plus~\citep{zhang2026world} by integrating it with $\pi_\text{0.5}$. As shown in Table~\ref{tab:robotwin}, R2A improves the average success rate from 56.8\% to 66.2\% and outperforms the base policy in all seven perturbation categories. The largest gain occurs under Robot Initial States, where success increases from 34.2\% to 75.9\%, followed by Light Conditions with a gain of 9.9\%. These results show that the benefits of R2A extend beyond single-arm settings to bimanual manipulation, further demonstrating its effectiveness under challenging OOD execution contexts.

\vspace{0.05in}
\textbf{Q2. How does R2A compare with existing OOD methods?}
We compare R2A with methods designed to address OOD conditions at deployment in LIBERO-10. VLS~\citep{liu2026vls} employs a VLM agent to detect object keypoints and generate reward functions for steering action generation, while SDN~\citep{nguyen2026sdn} uses negative references to steer actions away from spurious behaviors associated with OOD regions. We also adopt an in-context learning (ICL) baseline that retrieves the most aligned training observations and provides their textual descriptions as inference context. 

Table~\ref{tab:ood-baseline} reports success rates across seven perturbation categories. On average, R2A outperforms the other OOD baselines, with gains of 11.2\% over SDN for GR00T N1.7 and 2.5\% for $\pi_\text{0.5}$. While SDN performs strongly under Sensor Noise, it is less effective under Camera Viewpoints and Robot Initial States, where it can degrade the base policy. This is because SDN relies on object-centric negative references, which are informative when spurious behavior is caused by irrelevant visual cues, but become less reliable when the shift arises from the execution context, such as camera configuration or robot initialization. In these cases, the negative reference may not correspond to the source of the failure and can instead steer the policy toward an undesired action distribution. VLS exhibits a similar limitation under context shifts due to its object-centric steering, and further depends on the quality of reward functions generated by an external VLM agent, whose awareness of the deployment environment may be limited. In contrast, R2A retrieves motion primitives from training experiences according to the current execution context, providing positive action guidance that remains relevant under such shifts.

\vspace{0.1in}
\textbf{Q3. Does R2A improve robustness under real-world perturbations?} We evaluate R2A with GR00T N1.7 with Aloha robot (Sec.~\ref{sec:real-world_setup}) under perturbed robot initial states on Banana2Pot and Cube2Drawer.

As shown in Table~\ref{tab:real}, these perturbations substantially degrade the base policy and SDN. R2A is markedly more robust: it largely preserves in-distribution performance on Banana2Pot and improves Cube2Drawer success from 10\% to 45\%, exceeding SDN by 15\%. R2A also improves performance without perturbations, suggesting that retrieving relevant motion-primitive experience helps the policy recover from shifted initial states while supporting more reliable execution.

\input{table/real_component}

\vspace{-0.15in}
\paragraph{Ablation Study.}
\input{figure/latency}
\textbf{(a) Contributions of Components in R2A.}
We ablate the key components of R2A with GR00T N1.7 on LIBERO-Plus. Table~\ref{tab:component_r2a} replaces GNN Retrieval with directly selecting top-$k$ most aligned nodes and removes Action Harmonization by directly executing retrieved motion primitives.
Both components are essential. GNN Retrieval identifies relevant prior experiences, with direct selection reducing success from 77.1\% to 63.8\%. Action Harmonization is even more critical: directly replaying retrieved actions drops success to 1.2\%, highlighting the need to adapt prior motions to the current context. Finally, under Robot Initial States perturbations, Initial State Rectification (RIS) raises success from 49.9\% to 87.5\% by moving OOD initial states toward the training distribution. Overall, these findings show that R2A’s robustness arises from the complementary roles of retrieving relevant prior experience, adapting it to the current context, and reducing initial-state distribution shift

\textbf{(b) Inference Efficiency.}
Figure~\ref{fig:latency} reports the per-step inference latency and success rate of each method. R2A achieves the highest success rate while introducing only a modest latency overhead over the base policy, with an average per-step latency of 159ms compared to 138ms for the base model. In contrast, SDN requires external object detection and segmentation for object masking, introducing additional computational overhead. VLS further incurs substantially higher latency, as it queries an external VLM agent multiple times within each generation step, resulting in an inference latency exceeding 800ms. These results demonstrate that R2A provides an effective balance between task performance and inference efficiency.

%% file: table/main_table.tex
\definecolor{rTwoAccent}{HTML}{6245A0}
\definecolor{rTwoTint}{HTML}{F1EDF8}
\definecolor{rTwoGain}{HTML}{16804A}
\definecolor{rTwoLoss}{HTML}{C83E45}
\begin{table}[htbp]
\vspace{0.1in}
\caption{Success rates (\%) under seven LIBERO-Plus perturbation categories. }
\vspace{-0.1in}
\centering
\small
\setlength{\tabcolsep}{3.2pt}
\renewcommand{\arraystretch}{1.13}
\label{tab:main}
\resizebox{0.8\columnwidth}{!}{%
\begin{tabular}{l|ccccccc@{\hspace{12pt}}c}
\toprule
Model & Camera & Robot & Language & Lighting & Background & Noise & Layout & \textbf{Avg.} \\
\midrule
OpenVLA & 0.8 & 3.5 & 23.0 & 8.1 & 34.8 & 15.2 & 28.5 & 16.3 \\
OpenVLA-OFT & 56.4 & 31.9 & 79.5 & 88.7 & 93.3 & 75.8 & 74.2 & 71.4 \\
NORA & 2.2 & 37.0 & 65.1 & 45.7 & 58.6 & 12.8 & 62.1 & 40.5 \\
WorldVLA & 0.1 & 27.9 & 41.6 & 43.7 & 17.1 & 10.9 & 38.0 & 25.6 \\
UniVLA & 1.8 & 46.2 & 69.6 & 69.0 & 81.0 & 21.2 & 31.9 & 45.8 \\
$\pi_0$ & 13.8 & 6.0 & 58.8 & 85.5 & 81.4 & 79.0 & 68.9 & 56.2 \\
$\pi_0$-FAST & 65.1 & 21.6 & 61.0 & 73.2 & 73.2 & 74.4 & 68.8 & 62.5 \\
RIPT-VLA & 55.2 & 31.2 & 77.6 & 88.4 & 91.6 & 73.5 & 74.2 & 70.2 \\
VLA-JEPA & 63.3 & 67.1 & 85.4 & 95.6 & 93.6 & 66.3 & 85.1 & 79.5 \\
\midrule
GR00T N1.7 & 65.2 & 45.8 & 83.9 & 95.5 & 94.3 & 83.1 & 71.8 & 77.1 \\
\rowcolor{rTwoTint}
\quad + \textcolor{rTwoAccent}{\textbf{R2A}} & 80.5\textsuperscript{\textcolor{rTwoGain}{+15.3}} & 94.9\textsuperscript{\textcolor{rTwoGain}{+49.1}} & \textbf{86.6}\textsuperscript{\textcolor{rTwoGain}{+2.7}} & 97.1\textsuperscript{\textcolor{rTwoGain}{+1.6}} & 94.2\textsuperscript{\textcolor{rTwoLoss}{\ensuremath{-}0.1}} & 91.1\textsuperscript{\textcolor{rTwoGain}{+8.0}} & 78.3\textsuperscript{\textcolor{rTwoGain}{+6.5}} & \textcolor{rTwoAccent}{89.0}\textsuperscript{\textcolor{rTwoGain}{+11.9}} \\

StableVLA & 75.3 & 50.6 & 70.8 & 95.8 & 89.9 & 85.9 & 78.5 & 78.1 \\
\rowcolor{rTwoTint}
\quad + \textcolor{rTwoAccent}{\textbf{R2A}} & \textbf{85.7}\textsuperscript{\textcolor{rTwoGain}{+10.4}} & \textbf{98.6}\textsuperscript{\textcolor{rTwoGain}{+48.0}} & 75.0\textsuperscript{\textcolor{rTwoGain}{+4.2}} & \textbf{98.5}\textsuperscript{\textcolor{rTwoGain}{+2.7}} & 96.8\textsuperscript{\textcolor{rTwoGain}{+6.9}} & \textbf{92.3}\textsuperscript{\textcolor{rTwoGain}{+6.4}} & 79.7\textsuperscript{\textcolor{rTwoGain}{+1.2}} & \textcolor{rTwoAccent}{89.5}\textsuperscript{\textcolor{rTwoGain}{+11.4}} \\

$\pi_{0.5}$ & 72.0 & 75.3 & 85.8 & 97.1 & 95.6 & 88.6 & 87.0 & 85.9 \\
\rowcolor{rTwoTint}
\quad + \textcolor{rTwoAccent}{\textbf{R2A}} & 78.7\textsuperscript{\textcolor{rTwoGain}{+6.7}} & 97.2\textsuperscript{\textcolor{rTwoGain}{+21.9}} & \textbf{86.6}\textsuperscript{\textcolor{rTwoGain}{+0.8}} & 96.3\textsuperscript{\textcolor{rTwoLoss}{\ensuremath{-}0.8}} & \textbf{97.1}\textsuperscript{\textcolor{rTwoGain}{+1.5}} & 89.2\textsuperscript{\textcolor{rTwoGain}{+0.6}} & \textbf{87.7}\textsuperscript{\textcolor{rTwoGain}{+0.7}} & \textcolor{rTwoAccent}{\textbf{90.4}}\textsuperscript{\textcolor{rTwoGain}{+4.5}} \\
\bottomrule
\end{tabular}}
\vspace{0.15in}
\end{table}

%% file: table/ood_and_real.tex
\begin{figure}[htbp]
    \centering
    \vspace{-0.15in}
    \begin{minipage}[t]{0.64\textwidth}
        \vspace{0pt} %

        \definecolor{rTwoAccent}{HTML}{6245A0}
        \definecolor{rTwoTint}{HTML}{F1EDF8}
        \definecolor{rTwoGain}{HTML}{16804A}
        \definecolor{rTwoLoss}{HTML}{C83E45}

        \scriptsize %
        \setlength{\tabcolsep}{1.8pt}
        \renewcommand{\arraystretch}{1.15}

        \captionof{table}{Success rates (\%) under seven RoboTwin2.0-Plus perturbation categories.}
        \label{tab:robotwin}
        \vspace{-0.2cm}
        \begin{tabularx}{\linewidth}{
            @{}
            l|
            *{8}{>{\centering\arraybackslash}X}
            @{}
        }
        \toprule
        Method & Camera & Robot & Lang & Light & Back & Noise & Layout & \textbf{Avg.} \\
        \midrule
        $\pi_{0.5}$ &
57.5 &	34.2 &	71.7 &	55.5 &	67.0 &	60.9 &	51.1 &	56.8\\
        \rowcolor{rTwoTint}
        \quad + \textcolor{rTwoAccent}{\textbf{R2A}}
         & \textbf{59.3}\textsuperscript{\textcolor{rTwoGain}{+1.8}}
         & \textbf{75.9}\textsuperscript{\textcolor{rTwoGain}{+41.7}}
         & \textbf{74.3}\textsuperscript{\textcolor{rTwoGain}{+2.6}}
         & \textbf{65.4}\textsuperscript{\textcolor{rTwoGain}{+9.9}}
         & \textbf{68.6}\textsuperscript{\textcolor{rTwoGain}{+1.6}}
         & \textbf{64.5}\textsuperscript{\textcolor{rTwoGain}{+3.6}}
         & \textbf{55.7}\textsuperscript{\textcolor{rTwoGain}{+4.7}}
         & \textcolor{rTwoAccent}{\textbf{66.2}}\textsuperscript{\textcolor{rTwoGain}{+9.4}} \\
        \bottomrule
        \end{tabularx}

        \vspace{0.2cm} %

        \captionof{table}{Success rate (\%) on LIBERO-10 across different OOD methods. VLS employs a VLM agent for reward generation, SDN mitigates reliance on irrelevant OOD cues, and ICL leverages in-context learning with textual descriptions.
}
        \begin{tabularx}{\linewidth}{
            @{}
            l|
            *{8}{>{\centering\arraybackslash}X}
            @{}
        }
        \toprule
        Method & Camera & Robot & Lang & Light & Back & Noise & Layout & \textbf{Avg.} \\
        \midrule

        $\pi_{0.5}$
         & 50.6 & 73.3 & 88.8 & \textbf{95.6} & 93.4 & 75.5 & 89.1 & 80.9 \\

        \quad + VLS
         & 45.7 & 44.3 & 79.4 & 80.3 & 72.3 & 56.8 & 75.6 & 64.9 \\

        \quad + SDN
         & 48.9 & 73.5 & 90.6 & \textbf{95.6} & 92.7 & \textbf{95.8} & 89.7 & 83.8 \\

        \quad + ICL
         & 45.1 & 79.1 & \textbf{91.9} & 89.1 & 79.2 & 78.9 & 80.7 & 77.7 \\

        \rowcolor{rTwoTint}
        \quad + \textcolor{rTwoAccent}{\textbf{R2A}}
         & \textbf{55.1}\textsuperscript{\textcolor{rTwoGain}{+4.5}}
         & \textbf{96.9}\textsuperscript{\textcolor{rTwoGain}{+23.4}}
         & \textbf{91.9}\textsuperscript{\textcolor{rTwoGain}{+3.1}}
         & \textbf{95.6}\textsuperscript{\textcolor{gray}{+0.0}}
         & \textbf{95.2}\textsuperscript{\textcolor{rTwoGain}{+1.8}}
         & 78.0\textsuperscript{\textcolor{rTwoGain}{+2.5}}
         & \textbf{91.3}\textsuperscript{\textcolor{rTwoGain}{+2.2}}
         & \textcolor{rTwoAccent}{\textbf{86.3}}\textsuperscript{\textcolor{rTwoGain}{+5.4}} \\

        \midrule

        GR00T N1.7
         & 44.0 & 60.0 & 82.0 & 89.0 & 84.0 & 51.0 & 59.0 & 67.0 \\

        \quad + VLS
         & 38.1 & 30.9 & 66.8 & 77.9 & 73.2 & 50.2 & 64.1 & 57.3 \\

        \quad + SDN
         & 41.1 & 34.1 & 81.2 & 87.0 & 86.5 & \textbf{89.3} & 74.1 & 70.5 \\

        \quad + ICL
         & 50.8 & 60.1 & \textbf{89.3} & 81.0 & \textbf{91.3} & 77.5 & 71.8 & 74.5 \\

        \rowcolor{rTwoTint}
        \quad + \textcolor{rTwoAccent}{\textbf{R2A}}
         & \textbf{58.9}\textsuperscript{\textcolor{rTwoGain}{+14.9}}
         & \textbf{87.5}\textsuperscript{\textcolor{rTwoGain}{+27.5}}
         & 84.6\textsuperscript{\textcolor{rTwoGain}{+2.6}}
         & \textbf{94.5}\textsuperscript{\textcolor{rTwoGain}{+5.5}}
         & 88.9\textsuperscript{\textcolor{rTwoGain}{+4.9}}
         & 78.6\textsuperscript{\textcolor{rTwoGain}{+27.6}}
         & \textbf{81.4}\textsuperscript{\textcolor{rTwoGain}{+22.4}}
         & \textcolor{rTwoAccent}{\textbf{82.1}}\textsuperscript{\textcolor{rTwoGain}{+15.1}} \\

        \bottomrule
        \end{tabularx}

        \vspace{0.15cm}
        \label{tab:ood-baseline}
    \end{minipage}%
    \hfill%
    \begin{minipage}[t]{0.32\textwidth}
        \vspace{0pt} %
        \centering
        \includegraphics[width=\linewidth]{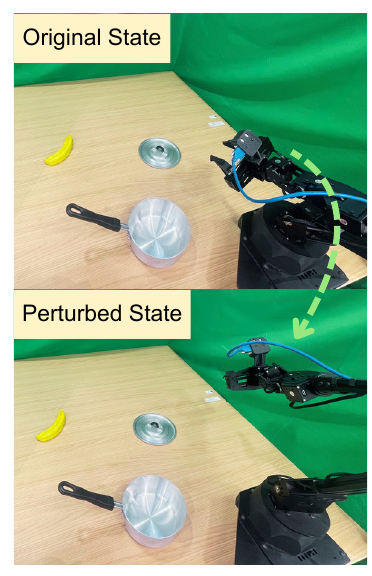}
        \captionof{figure}{Banana2Pot Task under Robot Initial States perturbation.}
        \label{fig:banana_state}
    \end{minipage}
\end{figure}

%% file: table/r2a-demo.tex
\begin{wrapfigure}[]{l}{0.47\textwidth}
    \centering
    \hspace{0.2in}
    \includegraphics[width=0.9\linewidth]{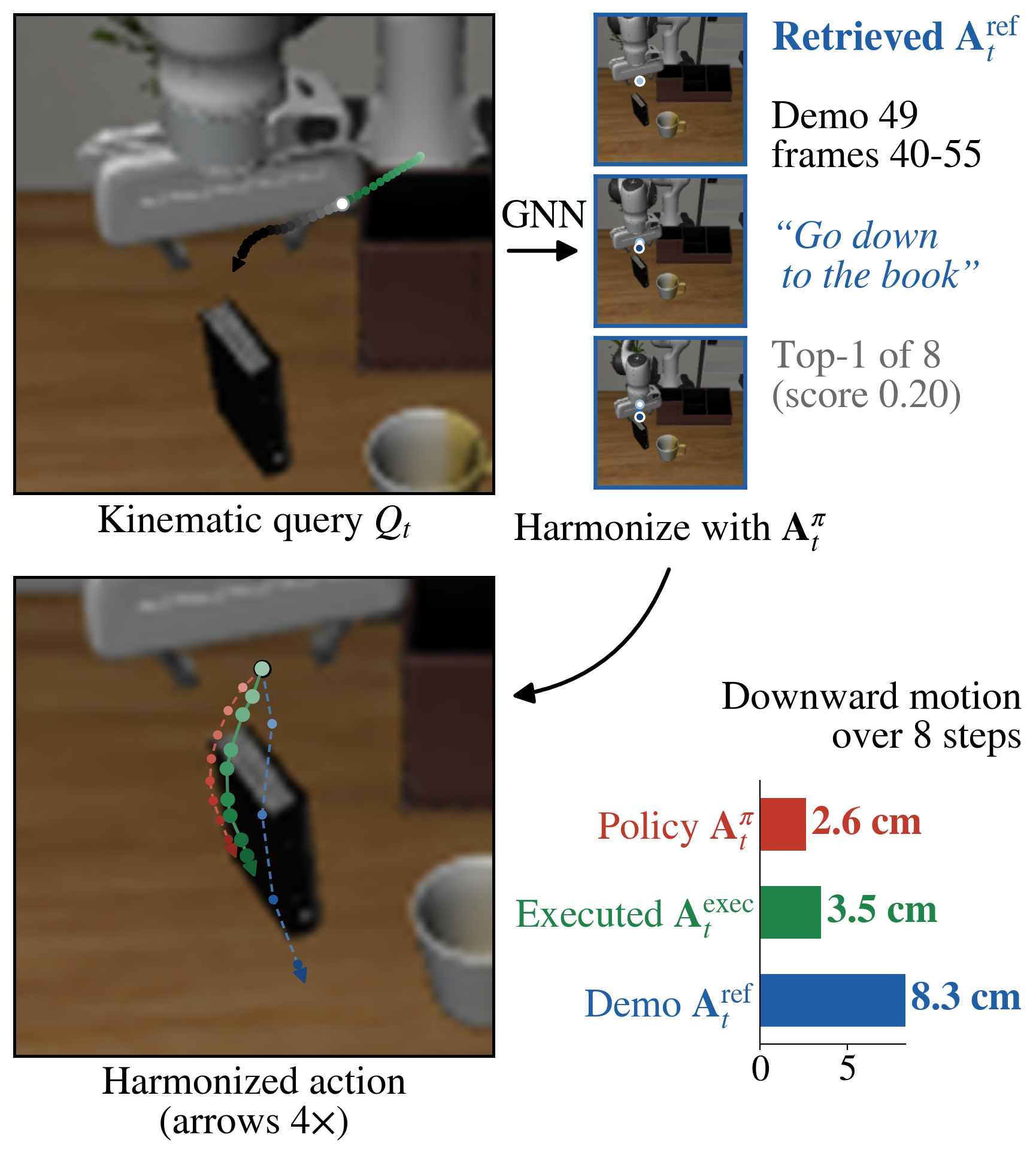}
    \vspace{-0.1in}
    \caption{R2A under Sensor Noise (GR00T N1.7). The query retrieves a demonstration going down to the book, and the executed action blends it with the policy. Background shown from a clean frame for readability.}
    \vspace{0.1in}
    \label{fig:r2a-demo}
\end{wrapfigure}

%% file: table/real_component.tex
\begin{table}[!t]
    \centering
    \vspace{-0.2in}
    \begin{minipage}[t]{0.48\textwidth}
        \centering
        \vspace{-0.18in}
        \caption{Performance under perturbation on robot initial states in real-world tasks using GROOT N1.7. We compare R2A against SDN, which steers diffusion noise to select the most stable actions.}
        \label{tab:real}
        \vspace{-0.25cm} %
        \small %
        \begin{tabular}{l|l|cc}
            \toprule
            Task & Method & Base & Robot \\
            \midrule
            \multirow{4}{*}{Banana2Pot}
            & GR00T N1.7 & 60.0 & 30.0 \\
            & + SDN & 75.0 & 55.0 \\
            & \cellcolor{rTwoTint}+ R2A 
              & \cellcolor{rTwoTint}\textbf{80.0}\textsuperscript{\textcolor{rTwoGain}{+20.0}}
              & \cellcolor{rTwoTint}\textbf{80.0}\textsuperscript{\textcolor{rTwoGain}{+50.0}} \\
            \midrule
            \multirow{4}{*}{Cube2Drawer}
            & GR00T N1.7 & 55.0 & 10.0 \\
            & + SDN & 70.0 & 30.0 \\
            & \cellcolor{rTwoTint}+ R2A 
              & \cellcolor{rTwoTint}\textbf{85.0}\textsuperscript{\textcolor{rTwoGain}{+30.0}}
              & \cellcolor{rTwoTint}\textbf{45.0}\textsuperscript{\textcolor{rTwoGain}{+35.0}} \\
            \bottomrule
        \end{tabular}
    \end{minipage}%
    \hspace{0.5cm} %
    \begin{minipage}[t]{0.45\textwidth}
        \centering
        \vspace{-0.15in}
        \caption{Contribution of R2A's component}
        \label{tab:component_r2a}
        \vspace{-0.25cm} %
        \small %
        \begin{tabular}{l|c}
            \toprule
            Method & Avg. SR (\%) \\
            \midrule
            \multicolumn{2}{c}{All axes} \\
            \midrule
            GR00T N1.7 & 77.1 \\
            R2A w/o GNN Retrieval & 63.8 \\
            R2A w/o Harmonization & 1.2 \\
            \textbf{R2A} w/o Ranking Loss & 80.7\\
            \textbf{R2A} & \textbf{89.0} \\
            \midrule
            \multicolumn{2}{c}{\begin{tabular}{@{}c@{}}Robot Initial States \end{tabular}} \\
            \midrule
            R2A w/o ISR & 49.9 \\
            R2A & 87.5 \\
            \bottomrule
        \end{tabular}
    \end{minipage}
    \vspace{0.15in}
\end{table}

%% file: figure/latency.tex
\begin{wrapfigure}{l}{0.6\columnwidth}
    \centering
    \vspace{-0.2in}
    \includegraphics[width=\linewidth]{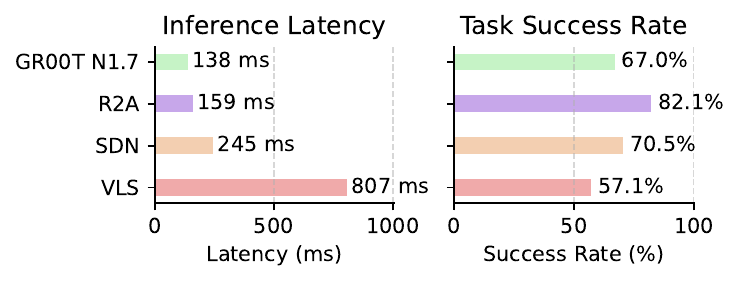} 
    \vspace{-0.2in}
    \caption{Comparison of Inference Latency and Task Success Rate across methods}
    \vspace{0.1in}
    \label{fig:latency}
\end{wrapfigure}

%% file: section/conclusion.tex
\section{Conclusion and Discussion}
In this work, we observed that using a kinematic graph is an effective strategy for overcoming out-of-distribution shifts that severely affect visual inputs. By rectifying out-of-distribution initial positions, properly retrieving geometrically aligned motion primitives, and adaptively harmonizing them with the policy's predictions, R2A achieves substantial improvements in execution robustness. 
The analysis separates localization stability, reference compatibility, and correction strength, giving sufficient conditions for pose-action improvement rather than a task-success guarantee.

\textbf{Limitations and Future Directions.}  R2A still presents several limitations. First, the retrieval index has no explicit scene geometry, so the method depends on the policy output to convey changes such as newly introduced obstacles. Second, it remains unclear how to extend this graph-based retrieval to explicit error-recovery settings~\citep{liu2026libero,chen2026resettle}, where the robot must autonomously recognize a physical task failure and execute a novel corrective sequence that does not exist in the prior demonstration graph. Finally, because the harmonization network is trained using policy and retrieval perturbations, it may fail to apply the necessary corrections if real-world visual shifts induce VLA prediction errors that fall outside this simulated distribution.

Building on these limitations, a promising direction for future work is to explore the validation and extension of R2A across broader architectural settings. First, (i) \textit{integrating R2A with few-shot or equivariant frameworks}~\citep{nguyen2026foca,ha2026equivla} offers an opportunity to investigate whether the offline spatiotemporal motion primitive graph can reliably anchor test-time execution when training demonstrations are severely restricted. Second, \textit{(ii) for non-Markovian environments} requiring distinct sub-task transitions, the framework could potentially incorporate object-centric memory states or semantic-graph planners~\citep{hanyu2026slotvla,chung2026rethinking,shi2026memoryvla,vo2026codegraphvlp}. This approach might allow the query-dependent GNN to condition retrieval on long-horizon temporal dependencies rather than strictly local states. Third, to address the absence of explicit scene geometry in highly cluttered environments, extending the retrieval mechanism to aggregate multi-view visual inputs~\citep{nguyen2026glovla,xu2026univiewvla,vo2025clutter} may help physically ground the retrieved motion primitives in robust 3D spatial arrangements. Finally, integrating R2A within post-training alignment pipelines~\citep{luu2026progress,fu2024robot} could facilitate the alignment of human reference motions with the retrieved geometric priors, encouraging the learned dimension-wise gating mechanism to harmonize actions toward both structural safety and human-preferred execution trajectories.

%% file: appendix/appendix.tex
\input{appendix/quantitative}
\input{appendix/qualitative}

\input{appendix/3_formulation}

%% file: appendix/quantitative.tex
\FloatBarrier   %

\section{Quantitative Results}
\label{sec:quantitative}
We evaluate GR00T N1.7 with and without R2A on ten manipulation tasks
from the LIBERO-Plus Long suite under seven perturbation axes: robot initial
configuration, sensor noise, camera pose, language, lighting, object layout,
and background. We use a paired evaluation protocol, where both policies are
tested on the same task, perturbation, and random seed. This allows us to
directly measure which episodes are fixed or broken by R2A.

\begin{figure}[H]
  \centering
  \includegraphics[width=\textwidth]{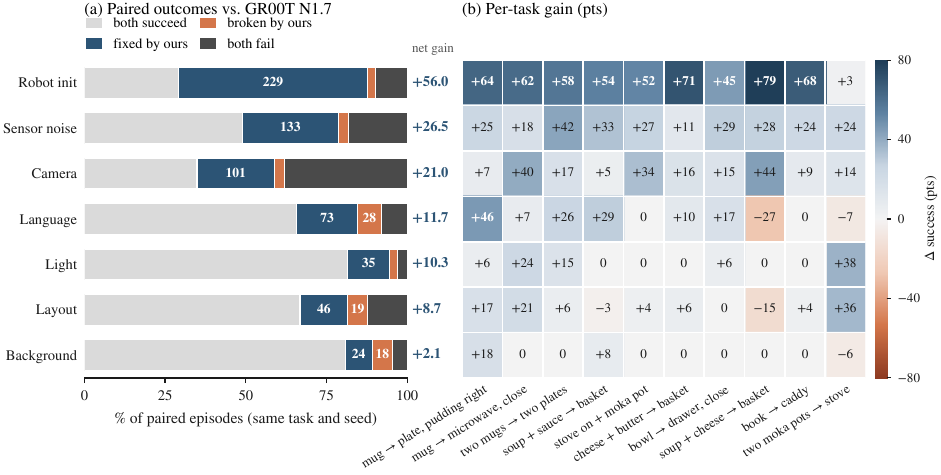}
  \caption{\textbf{Paired success outcomes.}
  (a) Paired outcomes for each perturbation axis. Each bar shows the fraction
  of episodes where both policies succeed, R2A fixes a GR00T N1.7 failure,
  R2A breaks a GR00T N1.7 success, or both policies fail. The value on the
  right is the net success-rate gain of R2A.
  (b) Per-task success-rate gain of R2A over GR00T N1.7 for each perturbation
  axis. Blue indicates improvement, orange indicates regression, and gray
  indicates no change.}
  \label{fig:quant-outcomes}
\end{figure}

\begin{figure}[!htbp]
  \centering
  \includegraphics[width=\textwidth]{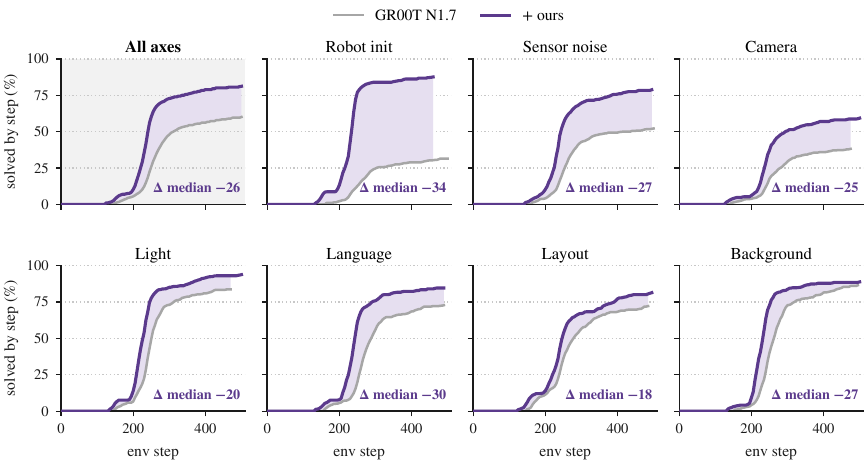}
  \caption{\textbf{Solving earlier.}
  Cumulative fraction of episodes solved by each environment step, pooled
  across all perturbation axes and shown separately for each axis.
  The shaded region shows the gap between R2A and GR00T N1.7.
  $\Delta$ median denotes the change in median solve step
  (R2A minus GR00T N1.7), where negative values indicate earlier success.}
  \label{fig:quant-solved-by-step}
\end{figure}

\begin{figure}[!htbp]
  \centering
  \includegraphics[width=\textwidth]{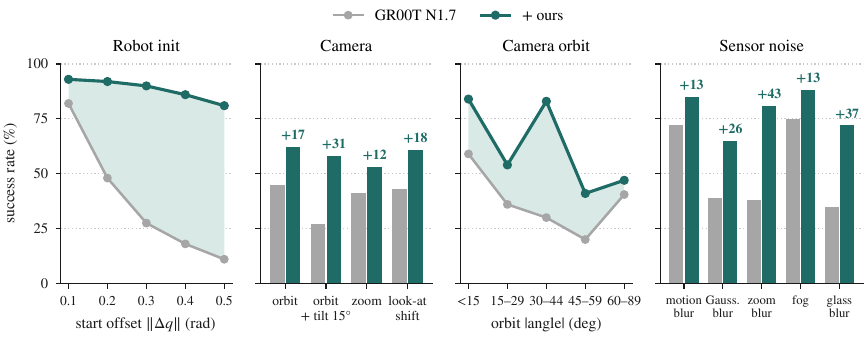}
  \caption{\textbf{Robustness under increasing perturbation.}
  Success rate of GR00T N1.7 and R2A as a function of robot start offset,
  camera perturbation type, camera orbit angle, and sensor-noise type.
  Numbers above the bars indicate the success-rate gain of R2A in percentage
  points.}
  \label{fig:quant-sweeps}
\end{figure}

We analyze the results from three perspectives: paired success outcomes
(Figure~\ref{fig:quant-outcomes}), time to success
(Figure~\ref{fig:quant-solved-by-step}), and robustness as perturbations become
stronger (Figure~\ref{fig:quant-sweeps}).

\paragraph{Paired outcomes.}
R2A improves the success rate across all seven perturbation axes
(Figure~\ref{fig:quant-outcomes}a). The largest gain occurs under robot
initialization, with a $+56.0$ point improvement, followed by sensor noise
($+26.5$) and camera perturbations ($+21.0$). These gains are driven by many
more fixed episodes than broken ones: R2A fixes $229$, $133$, and $101$
episodes on these three axes, respectively.

The gains are smaller for language, lighting, layout, and background.
Language and layout still show clear positive gains of $+11.7$ and $+8.7$
points, while background changes have only a small effect ($+2.1$ points).
Camera perturbations also contain a relatively large fraction of episodes
that neither policy solves, indicating that they remain challenging for both
methods.

\paragraph{Per-task breakdown.}
The improvements are distributed across most tasks
(Figure~\ref{fig:quant-outcomes}b). Under robot initialization, R2A improves nine
of the ten tasks by at least $45$ points, with the only small gain occurring
on \emph{two moka pots $\rightarrow$ stove} ($+3$ points).

Overall, only $5$ of the $70$ task--axis combinations show a negative change.
The largest regressions occur for
\emph{soup + cheese $\rightarrow$ basket} under language ($-27$) and layout
($-15$) perturbations. Most remaining task--axis combinations either improve
or remain unchanged.

\paragraph{Time to success.}
R2A not only solves more episodes but also tends to solve them earlier
(Figure~\ref{fig:quant-solved-by-step}). Across all perturbation axes, the final
fraction of solved episodes increases from $59.7\%$ for GR00T N1.7 to
$80.9\%$ with R2A, while the median solve step decreases by $26$ steps.

The same trend appears on every individual axis. The reduction in median
solve step ranges from $18$ steps for layout perturbations to $34$ steps for
robot initialization. In most cases, the curves begin to separate around the
main rise in successful episodes and remain separated for the rest of the
evaluation horizon, showing that R2A provides both higher success and earlier
completion.

\paragraph{Sensitivity to perturbation strength.}
Figure~\ref{fig:quant-sweeps} further shows how performance changes with different
perturbation strengths and types. As the robot start offset
$\|\Delta q\|$ increases from $0.1$ to $0.5$\,rad, the success rate of
GR00T N1.7 drops from $82\%$ to $11\%$. In comparison, R2A decreases only
from $93\%$ to $81\%$, widening the performance gap from $11$ to $70$
points.

R2A also improves performance across all tested camera perturbation types,
with gains ranging from $+12$ points for zoom to $+31$ points for orbit with
a $15^\circ$ tilt. When camera orbit perturbations are grouped by angle,
R2A remains above the baseline in every bin, although performance is not
strictly monotonic with the orbit angle.

For sensor noise, R2A again improves all tested conditions. The largest gains
appear under zoom blur ($+43$ points) and glass blur ($+37$ points), while
motion blur and fog show smaller gains of $+13$ points.

\FloatBarrier   %

%% file: appendix/qualitative.tex
\section{Qualitative Results}
\label{app:qualitative}

\subsection{LIBERO-Plus}
\begin{figure*}[p]
\centering
\includegraphics[width=\textwidth]{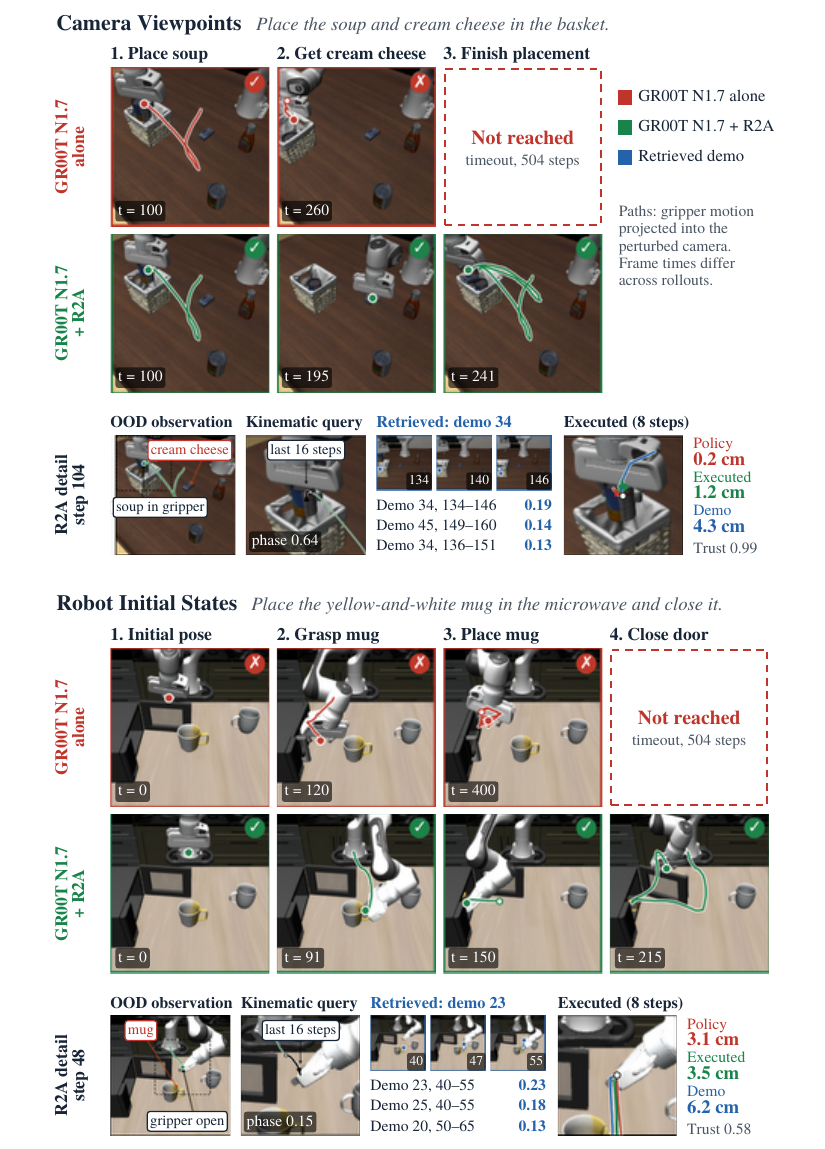}
\caption{Level-5 LIBERO-Long rollouts for Camera Viewpoints (top) and Robot Initial States (bottom). In each block, the first two rows compare GR00T N1.7 with and without R2A, and the third row details one R2A step. Camera Viewpoints: alone, the policy keeps pushing after placing the soup and never returns for the cream cheese; at step 104, the retrieved segment releases and rises (1.2 cm executed vs.\ 0.2 cm from the policy), and R2A finishes at step 242. Robot Initial States: alone, the policy heads for the microwave with an empty gripper; R2A first restores the start pose, then at step 48 retrieval moves the open gripper down toward the mug, and it finishes at step 225.}
\label{fig:qual}
\end{figure*}

Figure~\ref{fig:qual} compares representative frames from one level-5 LIBERO-Long episode for two perturbation categories, Camera Viewpoints and Robot Initial States, using the same GR00T N1.7 checkpoint with and without R2A. In each block, the first two rows distinguish the two methods and the columns indicate task stages. Frame times are shown explicitly and need not match across rollouts. Gripper trajectories are projected into the perturbed camera view.

The third row of each block details one R2A step. Its four panels show the OOD observation, the kinematic query, three frames from the highest-ranked demonstration segment together with the top-three retrieval scores, and the executed action compared with the policy prediction and retrieved demonstration. The query uses the last 16 steps of motion, including joint angles, velocities, and gripper state, together with the task instruction. The selected query steps are 104 (camera) and 48 (initial state). The initial-state query is shown after the initial pose correction; the two rollout prefixes are not assumed to be identical.

\subsection{Real-World Setup}
\label{sec:real-world_setup}
\textbf{Setup.}
We conduct real-world experiments on Mobile ALOHA~\citep{fu2024mobile}, equipped with two 6-DoF arms, a top-down RGB camera, and a wrist-mounted RGB camera, each capturing images at $640\times480$ resolution. For each task, we collect $150$ teleoperated demonstrations. We fine-tune GR00T~N1.7 from its 3B pretrained checkpoint using a batch size of $128$, a learning rate of $10^{-4}$, and $6000$ training steps, and use the resulting policy as our baseline. R2A is applied on top of this same fine-tuned GR00T~N1.7 policy. At inference time, the base policy uses $4$ denoising steps. We consider the following two real-world manipulation tasks:

\emph{Banana2Pot.} The robot picks up a banana, places it into a pot, and covers the pot with a lid. The banana, pot, and lid are randomly initialized within a $40\times60\ \mathrm{cm}$ workspace, with varying orientations across trials.

\emph{Cube2Drawer.} The robot opens a drawer, picks up a cube, places it inside, and closes the drawer. The cube and drawer are randomly initialized within a $40\times60\ \mathrm{cm}$ workspace. The cube yaw is randomized over the full $360^\circ$ range, while the drawer undergoes smaller orientation variations to keep its handle reachable.

The Robot Initial States perturbation places the arm outside the range of initial poses covered by the demonstrations while leaving the objects in place. Task-specific demonstrations are used to finetune the base policy and construct the motion primitive graph.

\begin{figure}[!ht]
\centering
\includegraphics[width=\textwidth]{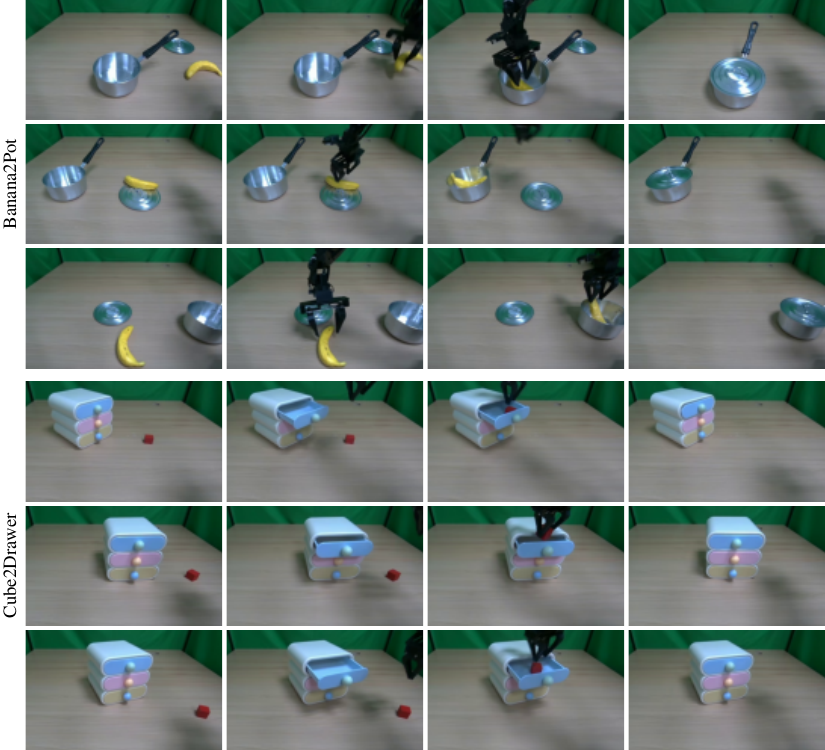}
\caption{Real-world demonstration samples for Banana2Pot (top) and Cube2Drawer (bottom). Each row shows one recorded episode (000, 075, and 149), and the columns go from the initial scene to the final scene. Frames come from the third-person camera and are cropped to reduce background.}
\label{fig:rw_data}
\end{figure}

%% file: appendix/implementation_detail.tex
\section{Experimental details}\label{app:experiments}

\subsection{Implementation details}\label{app:implement}

R2A is a test-time module attached to a frozen VLA policy. For each evaluation unit, the motion primitive graph and every auxiliary network are built from the demonstrations of that unit and from the frozen policy's actions replayed on those demonstrations. No base policy is updated, and no module sees an evaluation episode.

\textbf{Base policies and checkpoints.}
Every policy is used as released, without finetuning, and is queried with its own observations, instruction and action horizon (Table~\ref{tab:checkpoints}). GR00T N1.7 is served with the Isaac-GR00T code at tag \texttt{n1.7-release}, and $\pi_{0.5}$ with openpi. On LIBERO-Plus, GR00T N1.7 and StableVLA execute 8 actions per query and $\pi_{0.5}$ executes 5. On RoboTwin2.0-Plus, $\pi_{0.5}$ is the checkpoint released by the RoboTwin~2.0 authors, co-trained across RoboTwin~2.0 tasks for the AgileX ALOHA embodiment with joint-space actions; following their evaluation, the whole 50-step chunk of joint targets is executed before the next query. Instructions are embedded with a frozen \href{https://huggingface.co/Qwen/Qwen2.5-0.5B}{Qwen2.5-0.5B} for the task prior of the locator; the policies keep their own language encoders.

\begin{table}[ht]
\centering
\small
\caption{Base policy checkpoints, all used as released. Checkpoint names link to their sources.}
\label{tab:checkpoints}
\begin{tabular}{@{}lll@{}}
\toprule
Policy & Benchmark & Checkpoint \\
\midrule
GR00T N1.7  & LIBERO-Plus      & \href{https://huggingface.co/nvidia/GR00T-N1.7-LIBERO}{\texttt{nvidia/GR00T-N1.7-LIBERO}} (per suite) \\
$\pi_{0.5}$ & LIBERO-Plus      & \href{https://github.com/Physical-Intelligence/openpi}{\texttt{pi05\_libero}} (openpi) \\
StableVLA   & LIBERO-Plus      & \href{https://huggingface.co/DAGroup-PKU/StableVLA}{\texttt{DAGroup-PKU/StableVLA}} (per suite) \\
$\pi_{0.5}$ & RoboTwin2.0-Plus & \href{https://huggingface.co/datasets/TianxingChen/RoboTwin2.0}{\texttt{RoboTwin2.0} co-train, joint}$^{\dagger}$ \\
\bottomrule
\end{tabular}

\vspace{2pt}
\parbox{0.95\linewidth}{\footnotesize $^{\dagger}$\,File \path{cotrain_ckpt/Pi_05/RoboTwin-lerobot_v30-aloha_agilex-joint-0.tar.gz}.}
\end{table}

\textbf{Graph.}
Demonstrations are coarsened to one node every 5 frames, with edges to the 1, 2, 4, 8 and 16 nodes ahead along each demonstration. Segments are 16 frames long; the retriever shortlists $M=24$ segments and hands $K=8$ to the corrector. Every evaluation unit is built from 500 demonstrations (50 per task), giving 13k to 28k nodes on LIBERO-Plus and 8k to 10k nodes per arm on RoboTwin2.0-Plus.

\textbf{Model and training.}
R2A adds 0.37M trainable parameters per evaluation unit on top of the frozen policy (Table~\ref{tab:r2a_params}); the graph itself has no parameters. All widths and training settings are shared between the two benchmarks; only the input width of the policy-action projection changes with the policy's action chunk. The graph and the start-state alignment networks depend on the demonstrations alone and are shared across base policies; the locator reads $\mathbf A_t^\pi$, and the retriever and harmonization module are trained on its outputs, so these three are trained per base policy. Training reads only cached demonstrations and the frozen policy's replayed actions, and needs no robot rollouts.
\begin{table}[ht]
\centering
\small
\caption{R2A modules, sizes and training settings, identical for LIBERO-Plus and RoboTwin2.0-Plus. Adam throughout.}
\label{tab:r2a_params}
\begin{tabular}{@{}lp{5.2cm}rp{4.6cm}@{}}
\toprule
Module & Architecture & Params & Training \\
\midrule
Locator (GNN) & 3 message-passing layers, width 64, 16 relation types & 0.11M & 30 epochs \\
Retriever & 2-layer Transformer encoder (width 64, 8 heads) over the last 8 poses; segment key and action MLPs & 0.22M & \multirow{2}{4.6cm}{jointly, 30 epochs, lr $10^{-3}$, weight decay $10^{-5}$, 4{,}096 queries per epoch in batches of 256, 8 hard negatives per query; harmonization loss from epoch 5} \\
Harmonization & shape head (MLP, width 128) and trust head (MLP, width 64) & 5k & \\
Start-state alignment & score network (width 128, denoising score matching) and inverse-dynamics network (width 64) & 42k & 6{,}000 steps each \\
\midrule
Total & & 0.37M & \\
\bottomrule
\end{tabular}
\end{table}

\textbf{RoboTwin2.0-Plus deployment.}
Each arm is corrected on its own. One graph per arm is built from the group's demonstrations, in the arm's six commanded joints and its gripper state. The policy's chunk of joint targets is split into per-arm joint steps, each arm's steps are corrected against its graph, and the corrected steps are integrated back into joint targets. Three settings are specific to this benchmark and are the same for every group and axis. First, retrieval is joint over the two arms: a candidate segment for one arm is scored together with the segment of the other arm taken from the same demonstration and the same frames, so both arms follow one demonstration. Second, when an arm has not moved for three consecutive correction steps, and for longer than the retrieved demonstrations pause at the same phase, the corrector adds the displacement the demonstrations make from the current pose, provided the retrieved candidates agree on its direction (weighted cosine similarity at least 0.7). Third, after a gripper closes, the policy is queried every 10 action steps for the next 80 steps, so the object is carried under closed-loop visual feedback; the rest of the episode keeps the execution horizon of 50 steps.

\subsection{Evaluation protocol}\label{app:protocol}

\textbf{LIBERO-Plus.}
The evaluation unit is a LIBERO suite. One graph is built per suite from the suite's training demonstrations, so it holds the demonstrations of all tasks in the suite and localization and retrieval can move across tasks through the task prior and the \textsc{align} relation. Every perturbed episode of the released benchmark is run once under each of the seven perturbation categories; the number of episodes therefore varies by category. The comparison with OOD methods (Table~\ref{tab:ood-baseline}) and the ablations (Table~\ref{tab:component_r2a}) use the LIBERO-Long suite.

\textbf{RoboTwin2.0-Plus.}
The evaluation unit is a group of ten bimanual tasks, the same five alphabetical groups the RoboTwin~2.0 leaderboard uses. Each task is evaluated for 10 trials under each perturbation category.